\documentclass{article}
\usepackage{ijcai24}

\usepackage{times}
\usepackage{soul}
\usepackage{url}
\usepackage[hidelinks]{hyperref}
\usepackage[utf8]{inputenc}
\usepackage[small]{caption}
\usepackage{graphicx}
\usepackage{amsmath}
\usepackage{amsthm}
\usepackage{booktabs}
\usepackage{algorithm}
\usepackage{algorithmic}
\usepackage[switch]{lineno}
\usepackage{multirow}
\usepackage{array} 
\usepackage{amsfonts}
\usepackage{float}
\usepackage{makecell}

\usepackage{xcolor}

\title{DPGAN: A Dual-Path Generative Adversarial Network\\for Missing Data Imputation in Graphs}

\author{
Xindi Zheng$^1$
\and
Yuwei Wu$^2$\and
Yu Pan$^3$\and
Wanyu Lin$^1$$^\dagger$\and
Lei Ma$^{4,5}$\and
Jianjun Zhao$^3$
\affiliations
$^1$Hong Kong Polytechnic University\\
$^2$National University of Singapore\\
$^3$Kyushu University\\
$^4$University of Alberta\\
$^5$University of Tokyo\\
\emails
\{xin-di.zheng, wan-yu.lin\}@polyu.edu.hk,
yw.wu@nus.edu.sg,
panyu.ztj@gmail.com,
lei.ma@acm.org,
zhao@ait.kyushu-u.ac.jp
}

\begin{document}

\maketitle

\begin{abstract}
Missing data imputation poses a paramount challenge when dealing with graph data. Prior works typically are based on feature propagation or graph autoencoders to address this issue. However, these methods usually encounter the over-smoothing issue when dealing with missing data, as the graph neural network (GNN) modules are not explicitly designed for handling missing data. This paper proposes a novel framework, called Dual-Path Generative Adversarial Network (DPGAN), that can deal simultaneously with missing data and avoid over-smoothing problems. The crux of our work is that it admits both global and local representations of the input graph signal, which can capture the long-range dependencies. It is realized via our proposed generator, consisting of two key components, i.e., MLPUNet++ and GraphUNet++. Our generator is trained with a designated discriminator via an adversarial process. In particular, to avoid assessing the entire graph as did in the literature, our discriminator focuses on the local subgraph fidelity, thereby boosting the quality of the local imputation. The subgraph size is adjustable, allowing for control over the intensity of adversarial regularization. Comprehensive experiments across various benchmark datasets substantiate that DPGAN consistently rivals, if not outperforms, existing state-of-the-art imputation algorithms. The code is provided at \url{https://github.com/momoxia/DPGAN}.
\end{abstract}
{
\let\thefootnote\relax
\footnote{$\dagger$ denotes the corresponding author.}
}
\begin{figure}[t]
  \centering
  \includegraphics[width=\linewidth]{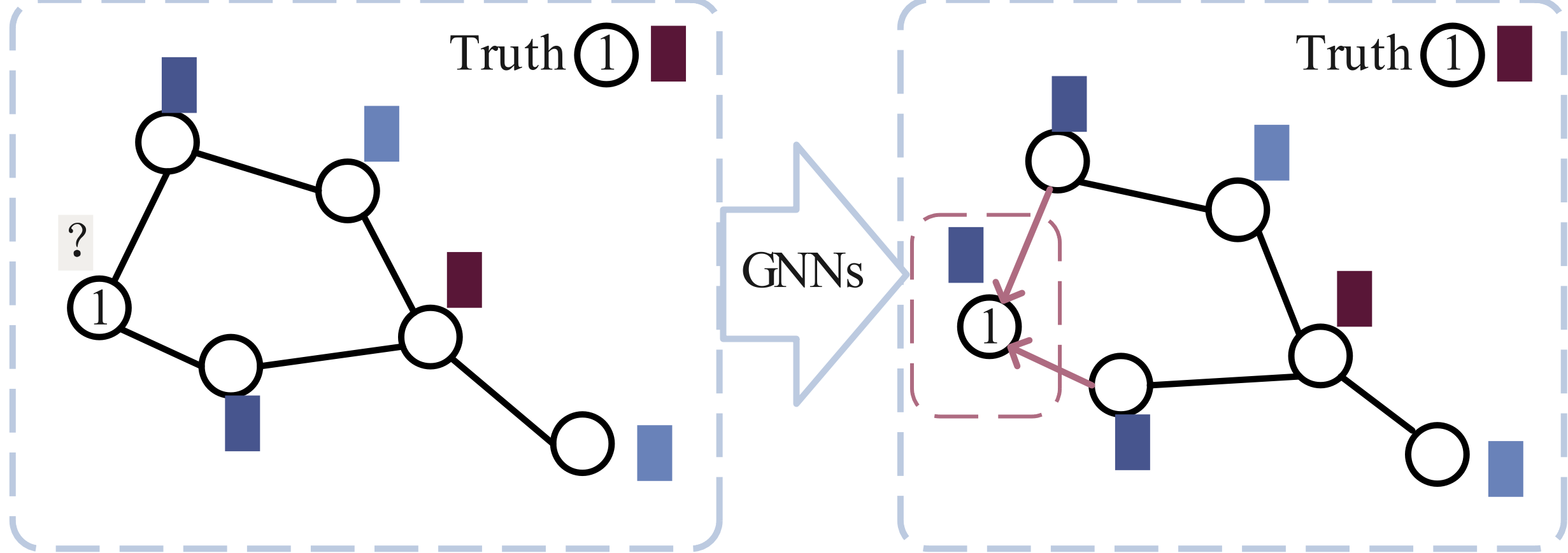}
  \caption{GNNs typically impute missing values by leveraging information from the neighboring nodes. In the graph, node 1 incorrectly imputed a missing value, which could be attributed to the influence of its neighboring nodes.}
  \label{problem}
\end{figure}

\section{Introduction}
Graph-structured data has been widely used to model various relationships for many real-world tasks such as traffic status forecasting \cite{li2017diffusion}, weather prediction \cite{ni2022weather}, and molecular classification \cite{wieder2020molecular}. However, missing graph attributes is a common issue due to the chaotic nature of the data collection process~\cite{yoon2016discovery}. For instance, data might be missing in a biochemical context because of the inherent difficulties of measuring or calculating quantitative molecular properties at an atomic scale(\cite{yomogida2012ambipolar}). 

Recent graph-learning-based methods have shown significant progress in addressing the above challenges. GCMF~\cite{taguchi2021graph} and FP~\cite{rossi2022fp} employ feature propagation for imputation purposes,  both of which highly depend on the propagation of local information, inducing sub-optimum solutions. GDN~\cite{li2020graph} develops a graph deconvolution operation for recovering graph features from smoothed representations. MEGAE~\cite{gao2023handling} attempts to address the spectral concentration problem by maximizing the graph's spectral entropy. These two works use autoencoders to compress node features into a latent space and then map them back to the original topology for imputation. While these autoencoders are promising, the generated results might be further improved through adversarial training, as shown in~\cite{spinelli2020missing}. Though GINN~\cite{spinelli2020missing} constructs a graph-based generator and discriminator, it neglects the over-smoothing issue and exhibits instability for high-quality graph feature generation.
%\xindi{Since the results acquire an additional objective of producing more realistic data to deceive the Discriminator.} \wanyu{not readable}. 

To address these gaps, our framework, consisting of a dual-path generator and a subgraph discriminator, is designed to augment the strengths and mitigate the weaknesses of traditional methods. We recognize that existing GNNs, while proficient in prediction, often falter in feature imputation. This leads to a competency mismatch between generators and discriminators in adversarial training, affecting overall efficacy. In this work, we introduce a novel dual-path model within graph feature generative adversarial networks (GANs) tailored to maintain stable performance across different rates of missing data. Our dual-path approach, comprising the GraphUnet++ and the MLPUnet++, offers a synergistic solution.

The first component of our dual-path generator is GraphUnet++, an innovative MLP-augmented GNN. Building upon the GraphUnet framework\cite{gao2019graphU}, it integrates GNNs for structural data processing with graph pooling layers for downsampling and a graph unpool layer for accurate reconstruction. This process efficiently reduces redundancy in feature and structural information by mapping larger graphs to a more compact graph latent space. A pivotal innovation in GraphUnet++ is the integration of Node-mix MLP layers following GNNs. These layers address the limitations of GNNs in capturing long-range dependencies, enhancing the global representation capabilities of GraphUnet++ for more effective feature imputation.

Complementing GraphUnet++, MLPUnet++ is our second path. Recognizing that traditional GNNs often struggle to capture global graph feature information, especially at lower missing data rates, MLPUnet++ utilizes Node-Mix MLP to facilitate node interactions and Feature-Mix MLP for feature interplay. Besides, the Node-Mix MLP condenses the node dimension to reduce redundancy between nodes. Integrating these two paths in a model ensemble framework allows each to focus on distinct aspects of graph feature learning, significantly enhancing model robustness across various missing data scenarios.

Finally, we propose a subgraph discriminator to further refine the generative process. This unique discriminator, operating on subgraphs of varying sizes, facilitates a balanced training process and enables the generator to undergo adversarial regularization at a suitable intensity, addressing the limitations of traditional single-output discriminators in balancing generator and discriminator capabilities.

Our contributions are as follows:
\begin{itemize}
\item We propose the MLP-augmented GraphUnet (GraphUneTt++). Traditional adversarial training with only GNNs faces the problem of oversmoothness, which weakens the generator's capability. To address this issue, we introduce Node-Mix MLP to solve the training problems of graph GANs. The introduction of Node-Mix MLP also enhances the model's ability to capture long-range dependencies to a certain extent.
\item We develop MLPUnet++, a new parallel model branch that further enhances the proposed method in terms of global feature representation, which is designed to capture long-range dependencies more stably. Experiments demonstrate that MLPUnet++ plays a supportive role at low missing rates.
\item We propose a subgraph adversarial training strategy, optimizing the performance of a graph-learning-based generator. This is the first GAN model tailored for graph features, applied to tackle the issue of graph feature imputation. The proposed framework achieves the best performance across various feature-missing rates and datasets.
\end{itemize}
\begin{figure*}[t]
\centering
\includegraphics[width=1\linewidth]{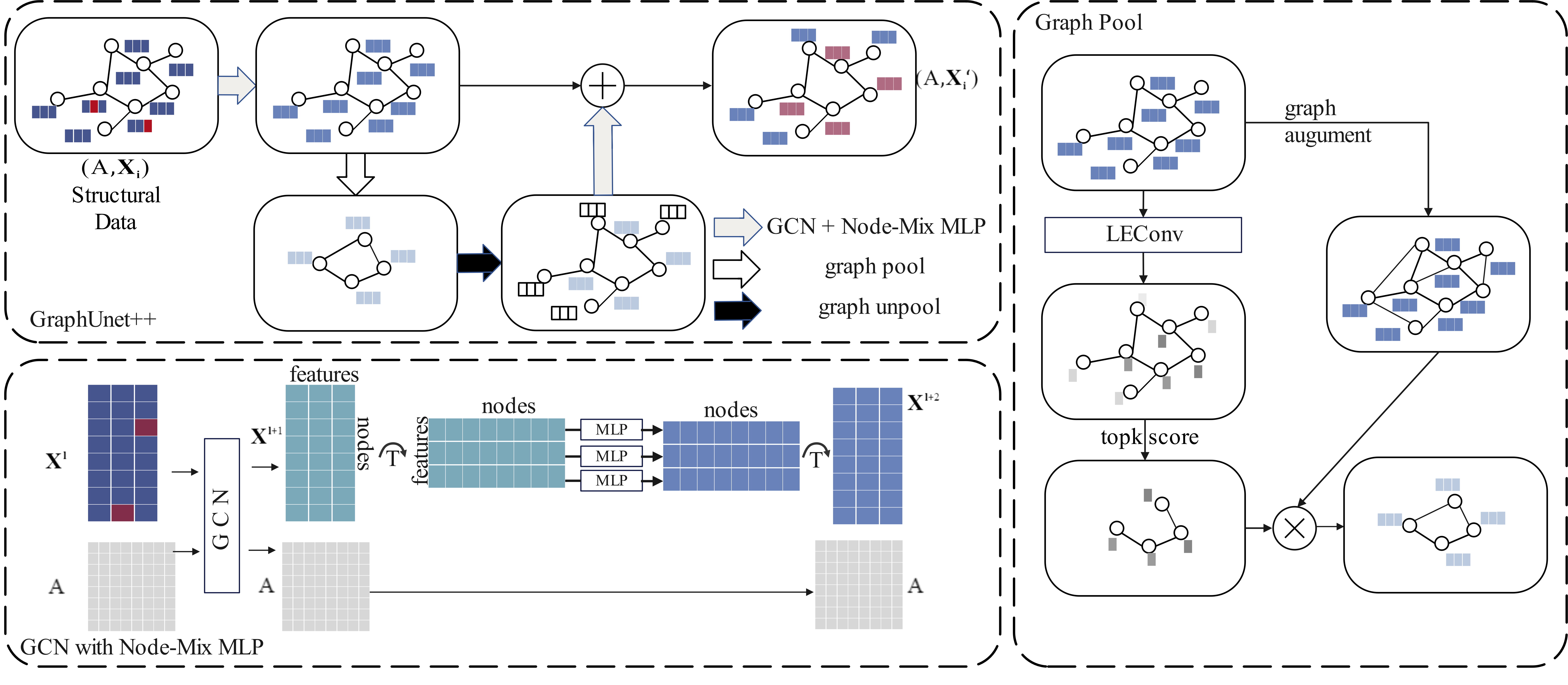}
\caption{Overview of GraphUnet++: Incorporating graph convolutional layers, node-mix MLP, graph pooling, and graph unpooling. The combination of GCN and node-mix MLP is utilized to extract both global and local representations of the graph. The node-mix MLP facilitates information exchange between nodes, with parameter sharing across all layers. Graph pooling is executed using LEConv to calculate the impact score for each node in clusters, followed by selecting top-k score nodes and the maintaining connectivity of subgraphs. Unpooling involves putting the feature of pooled graph back to its original corresponding nodes.}
\label{fig:GraphUnet}
\end{figure*}

\begin{figure*}[t]
\centering
\includegraphics[width=1\linewidth]{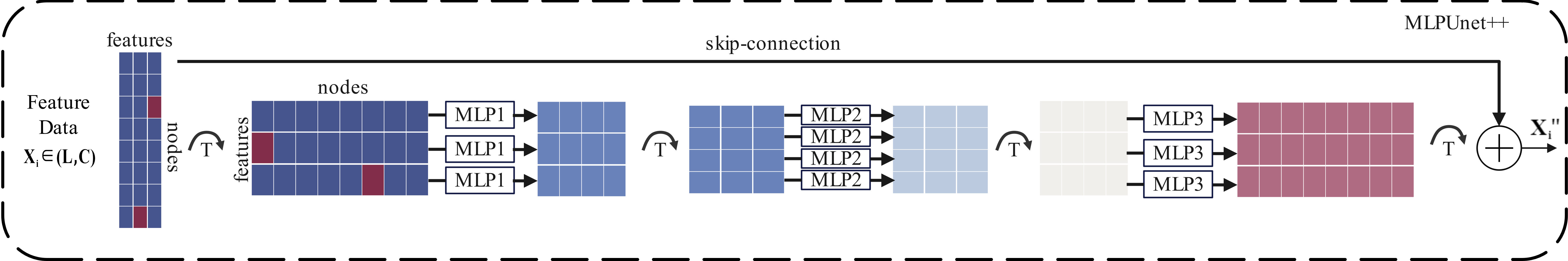}
\caption{MLPUnet++ Network consists of node-mix MLPs and feature-mix MLPs, each consisting of two fully connected layers and a LeakyReLU nonlinearity. Skip-connections are also included.}
\label{fig:MLPUnet}
\end{figure*}

\section{Related Work}
\subsection{Graph Autoencoder}
Since the introduction of Graph Neural Networks (GNNs) \cite{kipf2016gcn} and autoencoders (AEs), many studies \cite{kipf2016gcn} \cite{grover2019graphite} have used GNNs and AEs to encode to and decode from latent representations. Many autoencoder-based graph generation models, like Variational Auto-Encoders(VGAE), GraphVAE, and MolGAN. These models can be used in graph data imputation problems since they deal with both graph feature generation and structure generation. GraphUnet~\cite{gao2019graphU} proposed a U-net-like graph autoencoder, which we chose as our base graph autoencoder. GraphUnet introduces the conceptions of graph pooling for summarizing a graph to a small-size graph latent space and graph unpooling for recovering the small graph latent space to the original graph topology.
We use ASAPool~\cite{ranjan2020asap} instead of the graph pool layer in Graph-Unet.
\begin{align}
\begin{aligned}
& \mathbf{y} = \textrm{LEConv}(\mathbf{X}, \mathbf{A}), \ \mathbf{idx} = \mathrm{top}_k(\mathbf{y}) \\ 
& \mathbf{X}^{\prime} = (\mathbf{X} \odot \mathrm{tanh}(\mathbf{y}))_{\mathbf{idx}}, \  \mathbf{A}^{\prime} = \mathbf{A}_{\mathbf{idx},\mathbf{idx}}.
\end{aligned}
\end{align}
The formula of unpooling operation~\cite{gao2019graphU}:
\begin{equation}
\begin{aligned}
    {\mathbf{X}}^{\prime} = distribute(\mathbf{0}_{L\times C}, \mathbf{X}, \mathbf{idx}),
\end{aligned}
\end{equation}
where $\mathbf{idx} \in \mathbb{Z}^{k}$ contains indices of selected nodes in the corresponding pool layer that reduces the graph size from $L$ nodes to $k$ nodes. ${\mathbf{X}}^{\prime} \in \mathbb{R}^{k\times C}$ are the feature matrix of the current graph, and $\mathbf{0}_{L\times C}$, are the initially empty feature matrix for the new graph. $distribute(\mathbf{0}_{L\times C}, \mathbf{X}, \mathbf{idx})$ is the operation that distributes row vectors in ${\mathbf{X}}$ into $\mathbf{0}_{L\times C}$ feature matrix according to their corresponding indices stored in $\mathbf{idx}$.

\subsection{Adversarial Models}
Our method is motivated by the generative adversarial network (GAN) \cite{goodfellow2014generative}. GAN plays an adversarial game with two linked models: the generator G and the discriminator D. The discriminator discriminates if an input sample comes from the realistic data distribution or the generated data distribution. Simultaneously, the generator is trained to generate
the samples to convince the discriminator that the generated
samples come from the prior data distribution. G tries to generate samples to fool the discriminator, and D tries to differentiate samples correctly. 

\begin{equation}
\begin{aligned}
    & \mathop{\min}\limits_{G} \mathop{\max}\limits_{D} \mathcal{L}(D, G) = \\ 
    & \mathbb{E}_{x \sim p_{\text{data}}(x)} [\log D(x)] + \mathbb{E}_{z \sim p_z(z)} [\log (1 - D(G(z)))].
\end{aligned}
\end{equation}

To prevent undesired behavior such as mode collapse \cite{salimans2016improved} and to stabilize learning, we use Two Time-scale Update Rule (TTUR)\cite{heusel2017ttur} and improved WGAN \cite{gulrajani2017improved}, an alternative and more stable GAN model that minimizes a better-suited divergence. In our implementation, the formulation of adversarial loss is as follows:
\begin{equation}
\begin{aligned}
    & \mathcal{L}_{wgan-gp} = \mathbb{E}_{\mathbf{\tilde{x}} \sim \mathbb{P}_{gen}}\left[D\left(\tilde{\mathbf{x}}\right)\right] - \mathbb{E}_{\mathbf{x} \sim \mathbb{P}_{real}}\left[D\left(\mathbf{x}\right)\right] \\
    & \hspace{1.5cm} + \lambda_{GP}\mathbb{E}_{\mathbf{\hat{x}} \sim \mathbb{P}_{\hat{\mathbf{x}}}}\left[\left(||\nabla_{\hat{\mathbf{x}}}D\left(\mathbf{\hat{x}}\right)||_{2}-1\right)^{2}\right],
\end{aligned}
\end{equation}

where $\hat{\mathbf{x}}$ is a sampled linear combination between $x \sim p_{real}(x)$ and $\tilde{x} \sim p_{generated}(x)$ ,
thus $\hat{x}_{i} = \epsilon  x_i + (1-\epsilon) \tilde{x}_i $ with $ \epsilon \sim U(0, 1)$. The first
two terms measure the Wasserstein distance between real and fake samples; the last
term is the gradient penalty. As in the original paper, we set $\lambda_{GP} = 10$.

\section{Problem Definition}
In an undirected graph $G = (A; X)$, where $A \in \mathbb{R}^{N\times N}$ denotes the adjacency matrix and $X \in \mathbb{R}^{N\times D}$ denotes a comprehensive feature matrix. 
Here, $X_{ij}$ represents the attribute of the $i$-th node in the $j$-th feature dimension on the graph. In the context of graph attribute imputation, we define $R \in\{0; 1\}^{N\times D}$ as the mask matrix, where each element $R_{ij}$ equals 1 if $X_{ij}$ is observed and 0 otherwise.

This study aims to predict the missing graph attributes $X_{ij}$ where $R_{ij} = 0$. In essence, we aspire to construct a mapping function $f(\cdot)$ that generates the imputed data matrix $\tilde{X} \in \mathbb{R}^{N\times D}$, formally defined as follows:

\begin{equation}
\tilde{X} = f(X; R; A).
\end{equation}

\section{Proposed Algorithm}
\subsection{Framework}\
We aim to train a resilient generator for a given graph $G = (A; X)$. We accomplish this by employing an adversarial architecture along with a graph autoencoder and  MLPs autoencoder designed to process the entire graph directly and recover the missing rate. The overall workflow of DPGAN comprises two modules: the generator and the Subgraph discriminator.

\subsection{Generator}
The generator is comprised of an MLP (Multilayer Perceptron) autoencoder and a Graph autoencoder. Both leverage the graph structure $A$ and the node content $X$ as inputs to acquire a latent representation $Z$, and then they reconstruct the graph node features $X$ using $Z$. Each autoencoder implements skip connections to construct a U-net architecture. The Graph autoencoder, built upon the Graph U-net structure, is employed to restore the signal via the graph structure. In contrast, the MLP autoencoder is mainly concentrated on numerical feature fitting. The output of the generator is the weighted sum of the two autoencoders.\\
\textbf{Graph U-net++}
The input and output of data imputation problems differ in parts of features, and both are renderings of the same graph structure. We design the generator architecture around these considerations. Therefore, we consider the generator with skips.

Numerous preceding studies \cite{li2020graph} \cite{gao2023handling}  in this domain have employed an encoder-decoder network architecture. Within this framework, the input data is transformed into a latent space representation, from which features are subsequently reconstructed. A critical aspect of this approach is the necessity for information to traverse through all network layers, often resulting in a bottleneck. However, in many data imputation scenarios, a substantial portion of information is common to input and output. Therefore, it would be advantageous to facilitate the direct transfer of this shared information across the network, bypassing the bottleneck.

To facilitate the generator's access to observed information, we incorporate skip-connections into its design, inspired by the "Graph U-Net" architecture \cite{gao2019graphU}. While closely following the original Graph U-Net framework, our model diverges in one critical aspect: we introduce node-mix MLP layers following the GCN, which are specifically designed to capture the long-range dependencies of nodes. Additionally, we replace the standard pooling layer with an Adaptive Structure Aware Pooling (ASAP) layer \cite{ranjan2020asap}. This modification allows for more effective learning of hierarchical information. Our proposed architecture is named "Graph U-Net++". The overview of GraphUnet++ is shown in Figure \ref{fig:GraphUnet}, and the formulation of Graph U-Net++ is as follows:
\begin{equation}
\tilde{X}' = GraphUnet++(X; R; A).
\end{equation}
\textbf{MLP U-net++}
While Graph U-net can address the feature imputation problem, it can suffer from an oversmoothing issue, particularly when the network structure deepens, leading the network to lean towards low-frequency behaviors. To alleviate this problem, a model solely operating in the feature domain is needed to restore the lost high-frequency characteristics. This is achieved by employing a Multilayer Perceptron (MLP). The leading architecture substitutes the GCN part, as well as the pool and unpooling sections of the Graph U-net, with node-mix MLPs and feature-mixing MLPs respectively. The feature-mix MLPs are utilized, while the node-mix MLPs is employed to compress and expand the nodes' information and increase the information exchange between each node and all other nodes. Figure \ref{fig:MLPUnet} show the workflow of MLPUnet++. And also the formulation of MLPUnet++ is:
\begin{equation}
\tilde{X}'' = MLPUnet++(X; R; A).
\end{equation}

We use $\alpha$ to control the weights of the two channels. Therefore, the overall formulation for the generator is:

\begin{equation}
\begin{aligned}
&\tilde{X} = \alpha MLPUnet++(X; R; A) \\
&+ (1-\alpha)GraphUnet++(X; R; A)
\end{aligned}
\end{equation}

\subsection{Subgraph Discriminator}
The subgraph discriminator compels recovered data to conform to a prior distribution through an adversarial training module, which discriminates whether the current recovered data $\tilde{X} \in \mathbb{R}^{N\times D}$ originates from the fake distribution or the prior distribution.

It is well known that as the number of GCN layers increases, the recovered features tend to be dominated by low frequencies. In order to recover a high-frequency signal, it is sufficient to restrict our attention to the structure in the local graph. Therefore, we design a subgraph discriminator architecture – inspired by the patch discriminator from \cite{isola2017pix2pix} – that only penalizes the scale of the subgraph. The subgraph size is controlled by the number of graph pooling layers in the network. This discriminator attempts to classify whether each subgraph in a graph is real or fake.

\begin{figure}[H]
  \centering
  \includegraphics[width=\linewidth]{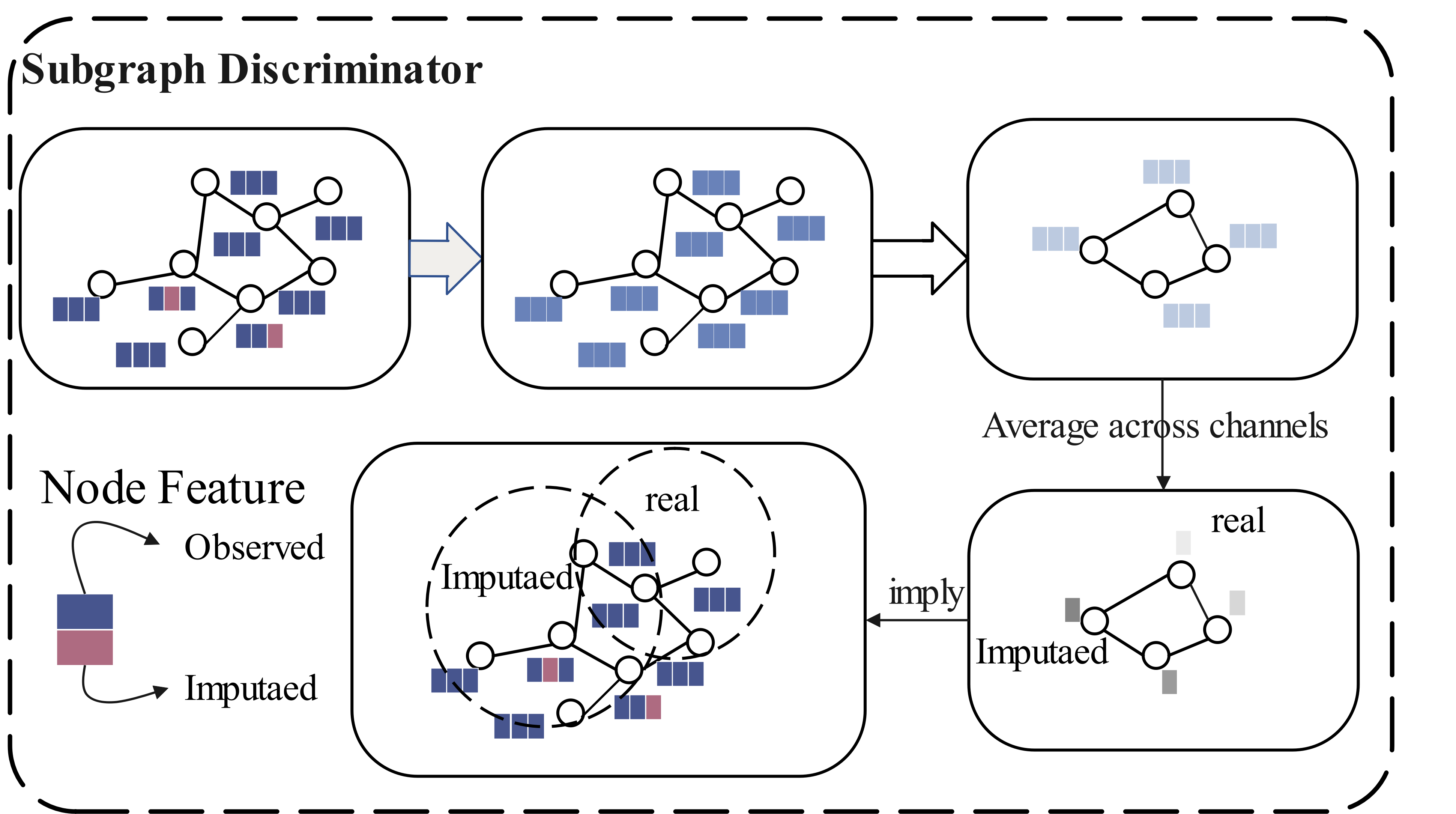}
\caption{Subgraph Discriminator: Following one graph convolution layer with a node-mix layer and one graph pooling layer, the graph size is reduced to 4 nodes. Each node represents the fidelity of the subgraph in its position.}
  \label{fig:discriminator}
\end{figure}

Furthermore, adversarial training necessitates that the capabilities of two modules, the generator and the discriminator, be on par. This is particularly crucial given that GNN-based generators often exhibit limited feature restoration ability. By allowing the discriminator to adjust its focus on subgraphs of varying sizes, we introduce a level of flexibility in the discriminator's capacity. One-hop subgraph discriminator, as shown in Figure~\ref{fig:discriminator}, will focus on the fidelity subgraph. The smaller number of hops implies more detailed observation, enabling a more nuanced analysis that is impossible with a single output per graph, which can easily overlook subtle details. Specifically, the zero hop subgraph discriminator or node discriminator means will output a fidelity score for each node. 

\begin{table*}[t]
    \centering
    \setlength{\tabcolsep}{6pt}
    \renewcommand{\arraystretch}{1.2}
   
    \begin{tabular}{c|ccccc|c}
    \Xhline{1.0pt}
    \multirow{2}{*}{Method} & ENZYMES  & QM9 & Synthie & FRANKE & FIRST\_DB & ENZYMES \\ \cline{2-6} 
    ~&  \multicolumn{5}{c|}{RMSE with 0.1 missing features} &ACC.\\
    \hline
    MEAN   & 0.0602  & 0.2983 & 0.2063 & 0.3891 & 0.1500 & 65.06\% \\
    KNN\cite{zhang2012knn}    & 0.0350  & 0.3058 & 0.1718 & 0.2010 & 0.1296 & 63.53\% \\
    SVD~\cite{troyanskaya2001svd}    & 0.0783  & 0.2524 & 0.1697 & 0.2766 & 0.1685 & 61.60\% \\
    MICE~\cite{van2011mice}   & 0.0292  & 0.1986 & 0.1899 & 0.1359 & 0.1036 & 64.46\% \\
    GAIN~\cite{yoon2018gain}   & 0.0300  & 0.1973 & 0.1649 & 0.1103 & 0.0909 & 64.42\% \\
    OT~\cite{muzellec2020missing}     & 0.0323  & 0.2003 & 0.1865 & 0.1116 & 0.0892 & 64.13\% \\
    MIRACLE~\cite{kyono2021miracle}& 0.0288  & 0.1846 & 0.1632 & 0.1196 & 0.0889 & 65.03\% \\
    GraphVAE~\cite{simonovsky2018graphvae}&0.0357  & 0.1579 & 0.1898 & 0.1099 & 0.1202 & 63.46\% \\
    MolGAN~\cite{de2018molgan} & 0.0326  & 0.1478 & 0.1864 & 0.1078 & 0.1379 & 64.16\% \\
    GRAPE~\cite{you2020grape}  & 0.0302  & 0.1869 & 0.1798 & 0.1069 & 0.0986 & 64.48\% \\
    GDN~\cite{li2020graph}    & 0.0268 & 0.1598 & 0.1764 & 0.1066 & 0.0869 & 65.57\%\\
    MEGAE~\cite{gao2023handling}  & \underline{0.0223} & \underline{0.1396} & \underline{0.1203} & \underline{0.0936} & \underline{0.0789} & \underline{66.27\%}  \\
    % \hline
    \textbf{DPGAN} (ours)  & \textbf{0.0193}   & \textbf{0.1011} & \textbf{0.1069} & \textbf{0.0908} & \textbf{0.0757}  & \textbf{67.34}\%     \\
    \hline
      & 0.0030 & 0.0385 & 0.0134 & 0.0028 & 0.0032 & 1.07\% \\
    Performance gain& $\big|$ & $\big|$  & $\big|$  & $\big|$  & $\big|$  & $\big|$  \\
    & 0.0409 & 0.2047 & 0.0994 & 0.2983 & 0.0928 & 5.74\% \\
    \Xhline{1.0pt}
    \end{tabular}
    \caption{RMSE results on five multi-graph datasets. After running five trials with random seed, we report the mean results in which the best method is bolded and the second best is underlined. Performance gains indicate the maximum (lower) and minimum (upper) gains of state-of-the-art (DPGAN) compared to other baselines. Note that we abbreviate ’FRANKENSTEIN’ and ’FIRSTMM\_DB’ to  ’FRANKE’ and ’FIRST\_DB’,respectively.}
    \label{tab:experiment1}
\end{table*}

\subsection{Optimization}
The overall training loss is defined as:
\begin{equation}
\mathcal{L} = \mathcal{L}_{WGAN-GP} + \lambda_{R} \mathcal{L}_\mathcal{R}
\end{equation}
where $ \mathcal{L}_{WGAN-GP}$ is the adversarial loss, $\mathcal{L}_\mathcal{R}$ is reconstruction loss and $\lambda_{R}$ is a hyperparameter to balance the regularization of
adversarial loss and reconstruction loss, we choose $\lambda_{R}$ from $\{1, 10, 100\}$.

In order to address the imputation precision, we introduce reconstruction Loss $\mathcal{L}_\mathcal{R}$. We follow the mean square loss for previous work \cite{li2020graph} \cite{gao2023handling}. In order to recover high-frequency signal\cite{isola2017pix2pix}, we also try L1 loss in the ablation study.
\begin{equation}
\mathcal{L}_\mathcal{R} = \Big\| (\mathbf{X} - \mathbf{\tilde{X}}) \odot (\mathbf{1}_{N \times D} - \mathbf{R}) \Big\|_2
\end{equation}

\section{Experiment}
In this section, we validate the performance of DPGAN
using a variety of datasets. We evaluate the effectiveness
of DPGAN on two categories of graph datasets:
\begin{itemize}
    \item \textbf{Imputation on multi-graph datasets.} We impute the missing graph attributes on multi-graph datasets, e.g., molecules and proteins. In addition, we report graph classification performance on graphs with imputed features.
    \item \textbf{Imputation on single-graph datasets.} We impute the missing values on single-graph datasets, e.g., social networks. We report RMSE on the graph with imputed features.
\end{itemize}

\subsection{Imputation on Multi-graph Datasets}
\subsubsection{Datasets} We conduct experiments on five benchmark
datasets\cite{morris2020tudataset} from different domains: 
(1)bioinformatics, i.e., ENZYMES \cite{schomburg2004enzyme}; 
(2) chemistry, i.e.,QM9 \cite{ramakrishnan2014quantum} and FIRSTMM\_DB \cite{neumann2013graph}; 
(3) computer vision, i.e., FRANKENSTEIN \cite{neumann2013graph}; 
(4) synthesis,i.e., Synthie\cite{morris2016faster}. Details of these datasets are summarized in Table \ref{tab:datasets}.
\begin{table}[h]
    \centering
    \renewcommand{\arraystretch}{1.2}
    \begin{tabular}{>{\centering\arraybackslash}m{25mm}>{\centering\arraybackslash}m{8mm}>{\centering\arraybackslash}m{8mm}>{\centering\arraybackslash}m{10mm}>{\centering\arraybackslash}m{10mm}}
    \Xhline{1.0pt}
    Datasets & \begin{tabular}[c]{@{}c@{}}Mean\\ Nodes\end{tabular} & \begin{tabular}[c]{@{}c@{}}Mean\\ Edges\end{tabular} & Features & \begin{tabular}[c]{@{}c@{}}Graph\\ Number\end{tabular} \\
    \hline
    ENZYMES & 33 & 62 & 18 & 600 \\
    QM9 & 18 & 19 & 16 & 1290\\
    Synthie & 95 & 173 & 15 & 400 \\
    FRANKENSTEIN & 17 & 18 & 780 & 4337 \\
    FIRSTMM\_DB & 1377 & 3074 & 1 & 41 \\
    \Xhline{1.0pt}
    \end{tabular}
    \caption{Summary of the experimental multi-graph datasets.}
    \label{tab:datasets}
\end{table}

\subsubsection{Baselines} We compare the performance of DPGAN
against baselines in three categories: (1) statistical imputation methods including MEAN, KNN~\cite{zhang2012knn} and
SVD~\cite{troyanskaya2001svd}; 
\begin{table*}[t]
    \centering
    \setlength{\tabcolsep}{2pt}
    \renewcommand{\arraystretch}{1.2}
    \begin{tabular}{c|ccccc|ccccc}
    \Xhline{1.0pt}
    \multirow{2}*{Method}&\multicolumn{5}{c}{Cora (RMSE)}&\multicolumn{5}{|c}{CiteSeer (RMSE)}\\ \cline{2-11} 
    ~& 0.1 Miss & 0.3 Miss & 0.5 Miss & 0.7 Miss & 0.99 Miss & 0.1 Miss & 0.3 Miss & 0.5 Miss & 0.7 Miss & 0.99 Miss\\
    \hline
    sRMGCNN\cite{monti2017geometric} & 0.1180 & 0.1187 & 0.1193 & 0.1643 & 0.1837 & 0.0693 & 0.0745 & 0.1137 & 0.1163 & 0.1750 \\
    GC-MC\cite{berg2017graph} & 0.0995 & 0.1089 & 0.1292 & 0.1571 & 0.2352 & 0.0599 & 0.0865 & 0.1032 & 0.1248 & 0.1892 \\
    GRAPE\cite{you2020grape} & 0.0975 & 0.1049 & 0.1256 & 0.1359 & 0.2274 & 0.0657 & 0.0930 & 0.1068 & 0.1295 & 0.1926 \\
    VGAE\cite{kipf2016vae} & 0.1105 & 0.1139 & 0.1616 & 0.2095 & 0.2892 & 0.0774 & 0.1060 & 0.1056 & 0.1350 & 0.2172 \\
    GDN\cite{li2020graph} & \underline{0.0946} & \underline{0.0964} & \underline{0.1085} & 0.1332 & 0.2037 & \underline{0.0599} & \underline{0.0895} & \underline{0.0893} & 0.1240 & 0.1784 \\
    MEGAE\cite{gao2023handling} & \textbf{0.0804} & \textbf{0.0849} & \textbf{0.0878} & \textbf{0.0941} & \underline{0.1208} & \textbf{0.0567} & \textbf{0.0621} & \textbf{0.0741} & \textbf{0.0938} & \underline{0.1408} \\
    % \hline
    \textbf{DPGAN} (ours)  & 0.1132 & 0.1124 & 0.1122 &\underline{0.1120}& \textbf{0.1119} & 0.0950 & 0.0943 & 0.0944 & \textbf{0.0938} & \textbf{0.0931} \\
    \hline
    Performance & -0.0328 & -0.0275 & -0.0244 & -0.0179 & 0.0089 & -0.0383 & -0.0322 & -0.0203 & 0.0000 & 0.0477\\
    gain        & $\big|$ & $\big|$  & $\big|$  & $\big|$  & $\big|$ & $\big|$ & $\big|$  & $\big|$  & $\big|$  & $\big|$ \\
    & 0.0048 & 0.0063 & 0.0494 & 0.0975 & 0.1773 & -0.0176 & 0.0117 & 0.0193 & 0.0412 & 0.1241 \\
    \Xhline{1.0pt}
    \end{tabular}
    \caption{Mean RMSE results of attribute imputation with different missing rates on Cora and CiteSeer. The best result is bolded and the second best is underline.}
    \label{tab:experiment2}
\end{table*}
(2) deep learning-based imputation models including MICE~\cite{van2011mice}, GAIN~\cite{yoon2018gain},
OT~\cite{muzellec2020missing} and MIRACLE~\cite{kyono2021miracle}; (3) graph learning-based models including GraphVAE~\cite{simonovsky2018graphvae}, MolGAN~\cite{de2018molgan}, GRAPE~\cite{you2020grape}, GDN~\cite{li2020graph}, MEGAE~\cite{gao2023handling}.

\subsubsection{Setup} We use a 70-10-20 train-validation-test split and
construct random missingness only on the test set. Each run
has a different dataset split and the mask for feature missingness. After running for five trials, we report the Root Mean Squared Error (RMSE) results for imputation on the test set. For all baselines in ENZYMES, we use a 2-layer GCN for downstream classification.
For hyperparameter tuning, we generally set the batch size to 128, except for the FIRST DB, which is set to 2. We select the number of hidden layers in the first GCN and MLP for both the Generator (G) and Discriminator (D) from the set \{128, 256, 512, 1024\}. Commonly, the combinations (256,256) and (1024,128) for the hidden layers of G and D, respectively, strike a good balance in their capabilities. For optimization, we use either Adam or SGD, and the initial alpha value is selected from \{0.5, 0.7, 0.9\}. As for learning rates, following the Two Time-Scale Update Rule\cite{heusel2017ttur}, the learning rate of D rate is set to 0.04, while the learning rate of G is chosen from \{0.01, 0.001, 0.0001\}.

\subsubsection{Results}
Table \ref{tab:experiment1} presents the experimental results with 10\% missing features. These results demonstrate that our approach consistently achieves the lowest RMSE across all five datasets, surpassing the current state-of-the-art (SOTA) by margins ranging from 2.99\% to 27.6\%. The improvement of imputation also benefits the downstream classification task.

\subsection{Imputation on Single-graph Datasets}
\subsubsection{Datasets} We assess DPGAN's efficiency on two widely recognized datasets, which are citation network datasets—Cora, Citeseer \cite{sen2008collective}.The detailed information of datasets is summaried in Table \ref{tab:single-datasets}.
\begin{table}[h]
    \centering
    \renewcommand{\arraystretch}{1.2}
    \begin{tabular}{>{\centering\arraybackslash}m{25mm}>{\centering\arraybackslash}m{8mm}>{\centering\arraybackslash}m{8mm}>{\centering\arraybackslash}m{10mm}>{\centering\arraybackslash}m{10mm}}
    \hline
    Datasets & \begin{tabular}[c]{@{}c@{}}Nodes\end{tabular} & \begin{tabular}[c]{@{}c@{}}Edges\end{tabular} & Features & \begin{tabular}[c]{@{}c@{}}Classes\end{tabular} \\
    \hline
    Cora & 2485 &5069 & 1433 & 7 \\
    QM9 & 2120 & 3679 & 3703 & 6\\
    \hline
    \end{tabular}
    \caption{Summary of the experimental single-graph datasets.}
    \label{tab:single-datasets}
\end{table}

\subsubsection{Comparison Methods} For imputation performance, we benchmark DPGAN against leading methods like sRMGCNN \cite{monti2017geometric}, GCMC \cite{berg2017graph}, GRAPE \cite{you2020grape}, VGAE \cite{kipf2016vae}, GDN\cite{li2020graph} and MEGAE\cite{gao2023handling}. 

\subsubsection{Experimental Framework} Our evaluation setup follows the previous work in \cite{gao2023handling}, primarily based on the approach by Kipf and Welling in 2017, where a standard dataset division is employed. Across all datasets, every iteration uses a distinct train-validation-test partition and a unique mask for random omissions in every feature dimension. After executing five trials, we present the average RMSE values for imputation.
\subsubsection{Results} 
The experimental results for a single graph dataset are shown in Table \ref{tab:experiment2}. Due to the vast number of points and distinct features in these datasets, and given the constraints on computational resources, MLPUnet could not participate. We employed a single-channel GraphUnet combined with a node-level subgraph discriminator. We opted for the node-level subgraph because using different levels of subgraph discriminators here would lead to leakage in the training set. The results show that the single-channel DPGAN performs best for single-graph tasks with high missing rates. However, without MLPUnet, achieving excellent results at low missing rates is challenging.
\subsection{Ablation Study}
\subsubsection{Objective Function and Generator Network}
We conduct an ablation study on the QM9 and ENZYMES datasets. After running five trials, we report the average results in Table \ref{tab:experiment3}. 
From the outcomes, it is evident that the dual-channel generator outperforms the single-channel generator. UsingeUtilize the L2 distance to construct the reconstruction loss is superior to the L1 distance, except for the ENZYME dataset with low missing rates. Generators with residual connections are more effective than the no-skip connection encoder-decoder structures. Train autoencoders with adversarial loss prove to be better than those trained solely with reconstruction loss.
\begin{table}[h]
    \centering
    \renewcommand{\arraystretch}{1.2}
    \begin{tabular}{c|cc}
    \hline
    Methods & QM9 & ENZYMES \\
    \hline
    $w/o$ Graph Unet & 0.1039 &  0.0224\\
    $w/o$ MLP Unet & 0.1074 & 0.0509\\
    
    $w/o$ Skip Connection & 0.1150& 0.0483\\
    $w/o$ WGAN$-gp$ &  0.1192& 0.0230\\
    \hline
    our method+$L_1$ & 0.1209&\textbf{0.0193}\\ 
    our method+$L_2$ & \textbf{0.1011} & 0.0221\\
    \hline
    \end{tabular}
    \caption{Mean RMSE results of attribute imputation with different Generator structures and Objective functions on QM9 and ENZYMES. We also test our model under different reconstruction loss metrics. The best result is bolded.}
    \label{tab:experiment3}
\end{table}

\subsubsection{From NodeDis to SubgraphDis to GraphDis}
We examine the impact of changing our discriminator's output subgraph by varying the discriminator's depth, i.e., the number of graph pooling layers. This ranges from a node discriminator (NodeDis), which has no graph pooling layer, to a full Graph discriminator (GraphDis), which gets the final score by adding a fully connected layer in the output of 2-hop subgraphDis. "2-hop SubgraphDis" implies that the discriminator has two graph pooling layers. Notably, in this paper, the pooling rate for all graph pool operations is 0.5.
The output of NodeDis retains the structure of the entire graph, and each node indicates the authenticity of that node. Hence, the authenticity of graph data is at the node level. This type of graph discriminator has been mentioned in \cite{spinelli2020missing}.
The output of GraphDis is a single authenticity score for an entire generated graph. This kind of single-value score structure has been seen in \cite{de2018molgan}.
Table~\ref{tab:dis} presents the effects of using discriminators at different levels by measuring the RMSE of the previous dataset. It is important to note that elsewhere in this paper, unless otherwise mentioned, all experiments utilize a subgraph discriminator with two graph pooling layers (2-hop SubgraphDis).

\begin{table}[t]
    \centering
    \renewcommand{\arraystretch}{1.2}
    \begin{tabular}{c|cc}
    \hline
    Methods & QM9 & ENZYMES\\ 
    \hline
    nodeDis & 0.1593 & \textbf{0.01932}\\
    1-hop subgraphDis &0.1027 & 0.02576\\
    2-hop subgraphDis & \textbf{0.1011} & 0.02567\\
    3-hop subgraphDis &0.1027 & 0.02691\\
    graphDis & 0.1581& 0.02706\\ 
    \hline
    \end{tabular}
    \caption{Mean RMSE results of attribute imputation with different sizes of discriminator on QM9  and ENZYMES. The best result is bolded.}
    \label{tab:dis}
\end{table}

It can be observed that using a subgraph discriminator, regardless of depth, can significantly enhance the effects of our GAN imputation data in the QM9 dataset.
Employing a 1-hop SubgraphDis is enough to achieve a commendable RMSE. The 2-hop SubgraphDis, albeit slightly, achieves better scores.
As we scale further, the poorer performance of NodeDis and GraphDis in QM9 is due to the fact that, during the training process, the generator struggles to counter a node-level discriminator; as such, a discriminator observes all nuances. On the other hand, a discriminator at the entire graph level can easily be bypassed, as the finer details get minimized in a single score.

Thus, achieving a balance between the generator and discriminator is challenging. This can be achieved by controlling the pooling layers to adapt effectively to adversarial training across various graph datasets.  Uniquely and notably in the ENZYME scenario, employing a 0-hop subgraph discriminator can yield enhanced results. 

\subsubsection{Imputation under various missing rates}
\begin{figure}[h]
  \centering
  \includegraphics[width=\linewidth]{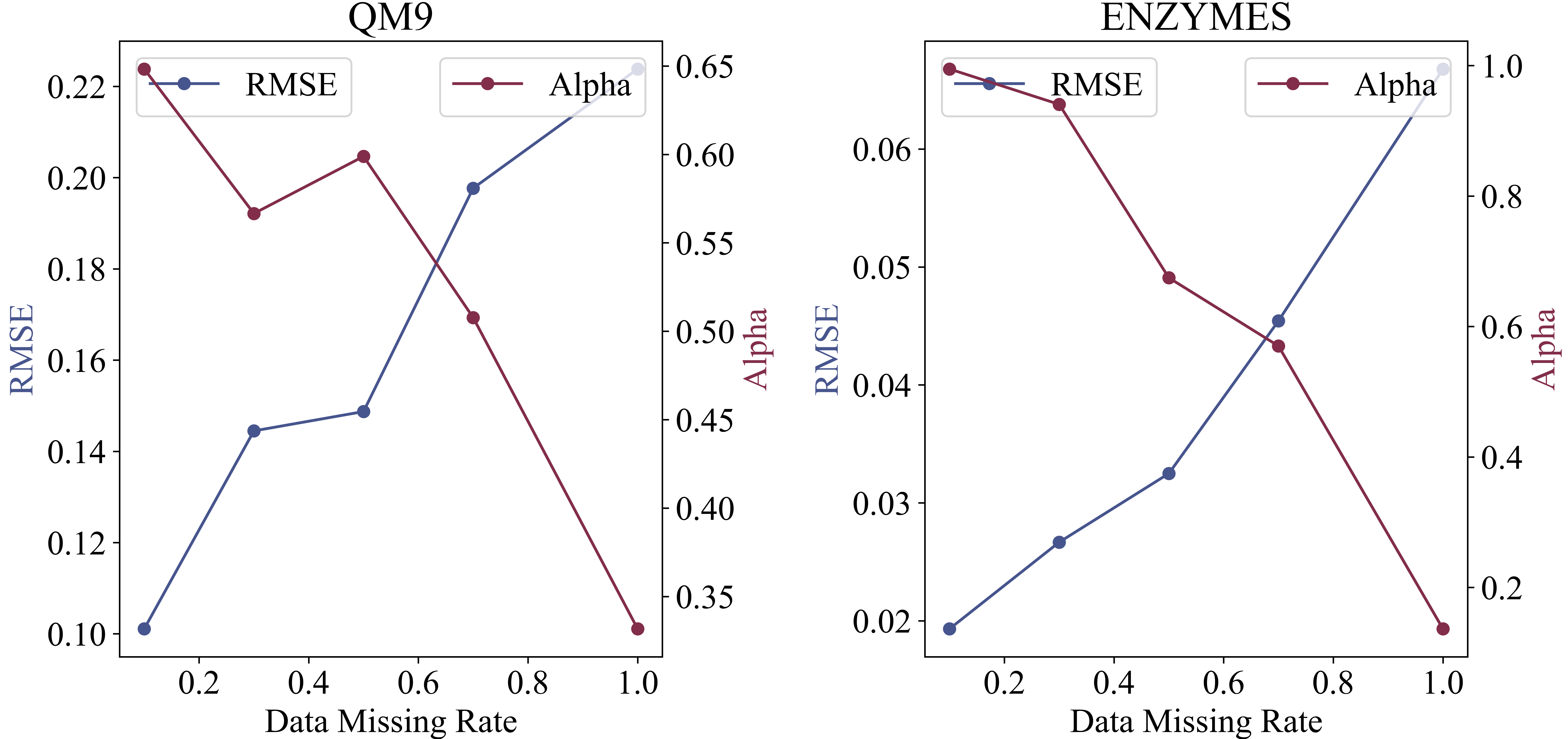}
  \caption{The imputation results under various missing rates. Our model is effective under various rates of data missing. Remarkably, it can reconstruct features relying solely on structure. As the rate of data missing varies, the final alpha value of the model also changes. A lower data missing rate results in a higher alpha value, indicating the dominant role of our designed MLPUnet.}
  \label{fig:missing}
\end{figure}

Figure \ref{fig:missing} shows that our proposed framework achieves high performance across various missing rates. Remarkably, it can effectively recover data even when all features are missing, a challenge that other algorithms have not addressed. Furthermore, as the missing rate decreases, the final $\alpha$ value increases, indicating that the specially designed MLPUnet++ plays a dominant role in the imputation process, particularly in scenarios with low missing rates. All the $\alpha$ values are initial at 0.5 and acquired after the training.

% \subsection{Computational Hardware}
% All models are trained with the following environment:
% \begin{itemize}
%     \item {Operating system: Ubuntu 22.04.3 LTS}
%     \item  {GPU: NVIDIA A100 Tensor Core GPU(40G)}
% \end{itemize}

\section{Conclusion}
This paper shows that dual-path autoencoders with subgraph adversarial regularization offer a promising approach for graph attribute imputation. The introduction of subgraph discriminators, which do not base their discrimination on a single value for evaluating a graph, enhances the training stability of graph-based GAN models, leading to better results in data imputation. The side-path Unet built on MLP further boosts the data imputation quality, although there are certain computational resource limitations.

%\section*{Ethical Statement}

%There are no ethical issues.

\bibliographystyle{named.bst}
\bibliography{ijcai24}

\end{document}